\documentclass[10pt,twocolumn,letterpaper]{article}

\usepackage{cvpr}
\usepackage{times}
\usepackage{epsfig}
\usepackage{graphicx}
\usepackage{amsmath}
\usepackage{amssymb}
\usepackage{soul}
\usepackage{caption}
\usepackage[font={small}]{caption}

\usepackage[pagebackref=true,breaklinks=true,letterpaper=true,colorlinks,bookmarks=false]{hyperref}

\cvprfinalcopy 

\def\cvprPaperID{4306} 

\ifcvprfinal\fi
\begin{document}

\title{Taming Adversarial Domain Transfer \\ with Structural Constraints for Image Enhancement}

\author{Elias Vansteenkiste and Patrick Kern\\
Brighter AI Technologies Gmbh\\
Torstrasse 177, Berlin\\
{\tt\small \{elias, patrick\}@brighter.ai}
}

\makeatletter
\let\@oldmaketitle\@maketitle
\renewcommand{\@maketitle}{\@oldmaketitle
  \includegraphics[width=\linewidth, trim={0cm 8cm 12cm 0}, clip]
    {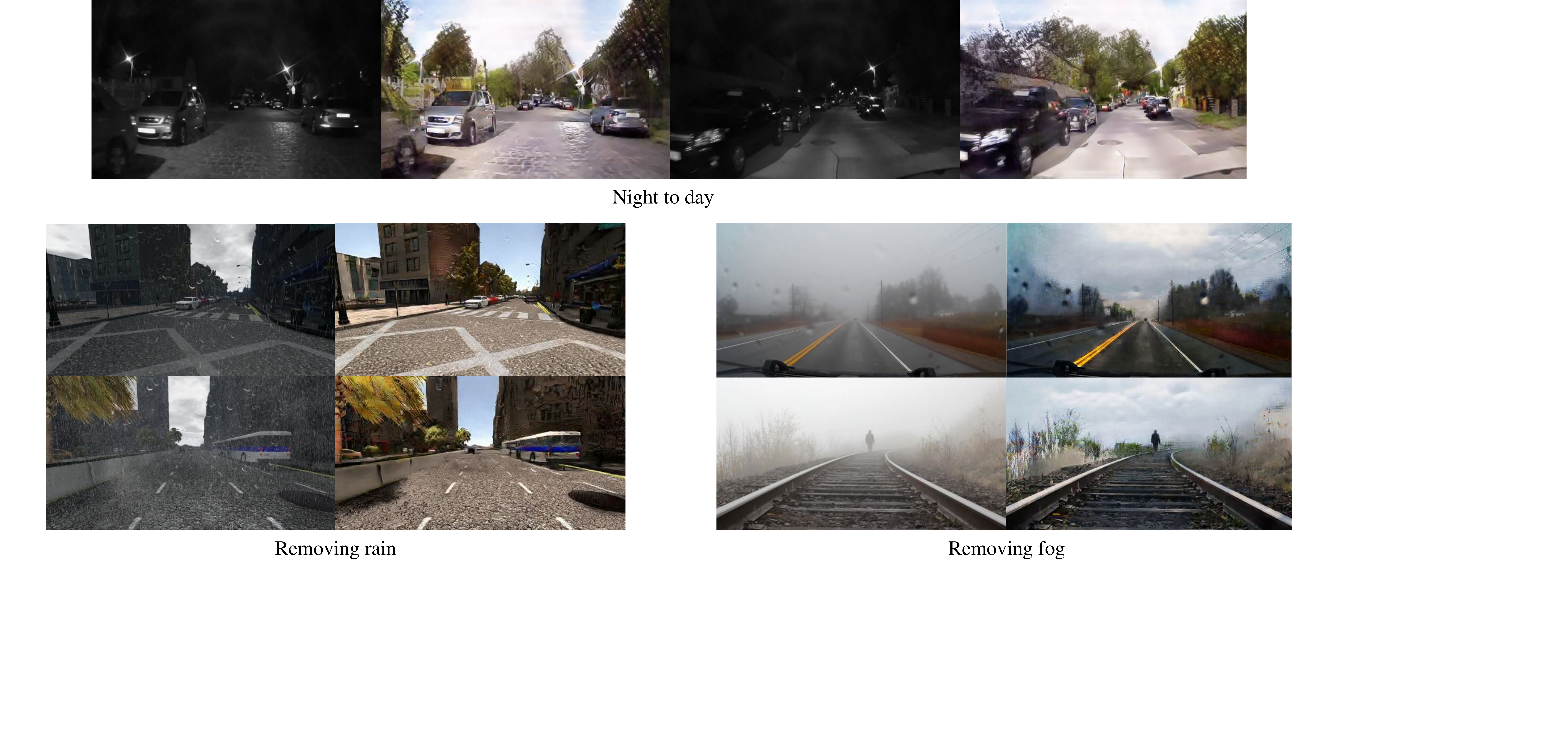}
  \captionof{figure}{\label{fig:front}Our domain transfer techniques applied to the night-to-day, removing rain and removing fog applications}
  \bigskip \smallskip}
\makeatother

\maketitle

\begin{abstract}
\vspace{-11px}
The goal of this work is to improve images of traffic scenes that are degraded by natural causes such as fog, rain and limited visibility during the night. For these applications, it is next to impossible to get pixel perfect pairs of the same scene, with and without the degrading conditions. This makes it unsuitable for conventional supervised learning approaches, however, it is easy to collect unpaired images of the scenes in a perfect and in a degraded condition. To enhance the images taken in a poor visibility condition, domain transfer models can be trained to transform an image from the degraded to the clear domain. A well-known concept for unsupervised domain transfer are cycle-consistent generative adversarial models. Unfortunately, the resulting generators often change the structure of the scene. This causes an undesirable change in the semantics. We propose three ways to cope with this problem depending on the type of degradation.

\end{abstract}

\section{Introduction}

Generative models can be used to produce new samples from the distribution represented by a given dataset. 
There are three important approaches to generative modelling: Variational Auto Encoders (VAE) \cite{vae}, Autoregressive (AR) models~\cite{pixelcnn} and Generative Adversarial Networks (GAN)~\cite{gans}. The latter is a basic building block in our method.
GANs consist of two parts, a generator G and a discriminator D. During training, a vector with values randomly sampled from a normal distribution is fed to the generator network. The generator outputs an image which should be visually indistinguishable from the samples in the dataset. To evaluate this, the discriminator outputs a value that indicates how real a sample looks. Both networks are trained simultaneously. While the objective for the generator is to fool the discriminator, the objective for the discriminator is to distinguish between real and generated samples. 

Generative adversarial training is also commonly used to transform images from one domain to another. 
The input to the transformation network is an image of one domain. The output of the transformation network should achieve two goals. The transformed image must be indistinguishable from the real samples in the target domain and the transformation should produce semantically correct mappings.

There are several methods for applying adversarial training for a supervised transformation task. An example is superresolution in \cite{superresolution}. The resolution of the input image is increased by the transformation. The objective function is defined by a weighted sum of an adversarial loss and a supervised loss. 
A second example is transforming satellite photos to maps and segmentation maps of street scenes to photorealistic renders in~\cite{pixtopix}. In this method the discriminator looks at pairs of images of both domains. It needs to distinguish between pairs with real images from both domains and pairs with a real image from the source domain and a generated image from the target domain.

Both these methods are characterised by the fact that the models are trained in a supervised manner because corresponding pairs of images from both domains are available. 

For unsupervised domain transfer the authors in~\cite{discogan} and~\cite{cyclegan} proposed to train two transformation and two discriminative networks simultaneously. 
A transformation network is built for each direction, an A2B and a B2A transformer, even though only the A2B transformation is required for some applications. The goal for each discriminator is to distinguish between the real and transformed images for their respective domain.
The novel idea is that by transforming an image from one domain to the other and then back it should result in a good reconstruction of the original image. To enforce this, the authors introduced a cycle consistency loss. The total loss is the sum of the generator, discriminator and cycle consistency loss $L_{\text{cyc}}$.

\newcommand{\norm}[1]{\left\lVert #1 \right\rVert}

\begin{gather} 
L = L_G^{A2B} + L_G^{B2A} + L_D^A + L_D^B + L_{\text{cyc}}\\ 
L_{\text{cyc}} = \lambda_{\text{cyc}}^{A} \cdot \norm{a - a_r}_{1} + \lambda_{\text{cyc}}^{B} \cdot \norm{b - b_r}_{1}
\end{gather}
with $\lambda_{\text{cyc}}^{A}$ and $\lambda_{\text{cyc}}^{B}$, the weights for balancing the consistency loss to the discriminator and generator losses.
In~\cite{cyclegan} the authors use an L1 distance between the original $(a,b)$ and the reconstructed image $(a_r, b_r)$ after a complete cycle, with $a\!\in\!A$ and $b\!\in\!B$.
Other unsupervised domain transfer approaches focus on small domain shifts without high level changes or geometric variations~\cite{bousmalis2016unsupervised}.

There are three important applications covered in this work, as illustrated in Figure~\ref{fig:front}. In all of them, the goal is to enhance the images of traffic scenes that are degraded by natural causes. The first application is clearing up rainy traffic scenes, which encompasses removing raindrops and making the overall scene brighter. The second application is transforming foggy traffic scenes. The fine details of the foreground objects should be reconstructed and give a hint of what is obfuscated by the fog in the distance.
In the third application the goal is to transform night near-infrared traffic imagery to clear bright daylight images. 
Since a common denominator of these problems is that it is very hard to collect corresponding paired images from both domains, supervised techniques can not be applied. It is however relatively easy to obtain datasets with unpaired images with degraded visibility and images taken in perfect conditions.

For all these applications, it is of prime importance that the semantics of the traffic situation do not change when the image is enhanced. Cars, pedestrians, buildings and trees need to be in the same location and retain their original shape. This is a problem with the current domain transfer approaches. If the transformers are trained with the cycle consistent adversarial method, the original semantic structure can completely change after transformation. The results are shown in Figure~\ref{fig:qualcomp_original_edge} and in more detail in Section~\ref{sec:add}. 

We propose three techniques to retain the structure and semantics of the traffic scene.
In the first two techniques, we explicitly force that commonalities between domains are retained during translation. 

If the domains are close to each other and the degradation has small features, like noise, then the perception should be the same in both domains~(Section~\ref{sec:perc}). Optimising for perception loss during training can guide the transformation network to keep the important features in the image. In case the domains are further apart in terms of color mapping, this does not work as well anymore. In that case forcing the transformer to keep the edges of the original image can provide good results as can be seen in Figure~\ref{fig:qualcomp_original_edge} and in~Section~\ref{sec:edge}.
In the last technique, domain knowledge is exploited by generating additional information that the discriminator network can use to make a better prediction. Even if it is only for one of the domains, it can help the whole cycle. In the night-to-day vision application, we use a pre-trained traffic scene segmentation network. The segmentation network is trained on images of traffic scenes during the day. For the daylight domain, segmentation maps are generated for both the transformed images and the original images and fed to the discriminator~(Section~\ref{sec:ss}).
Improvements to the transformation network architecture are described in Section~\ref{sec:arch}. In the next we introduce the three benchmark applications.


\begin{figure}[t]
\begin{center}
  \includegraphics[width=\linewidth, trim={0 8cm 4cm 0cm}, clip]
    {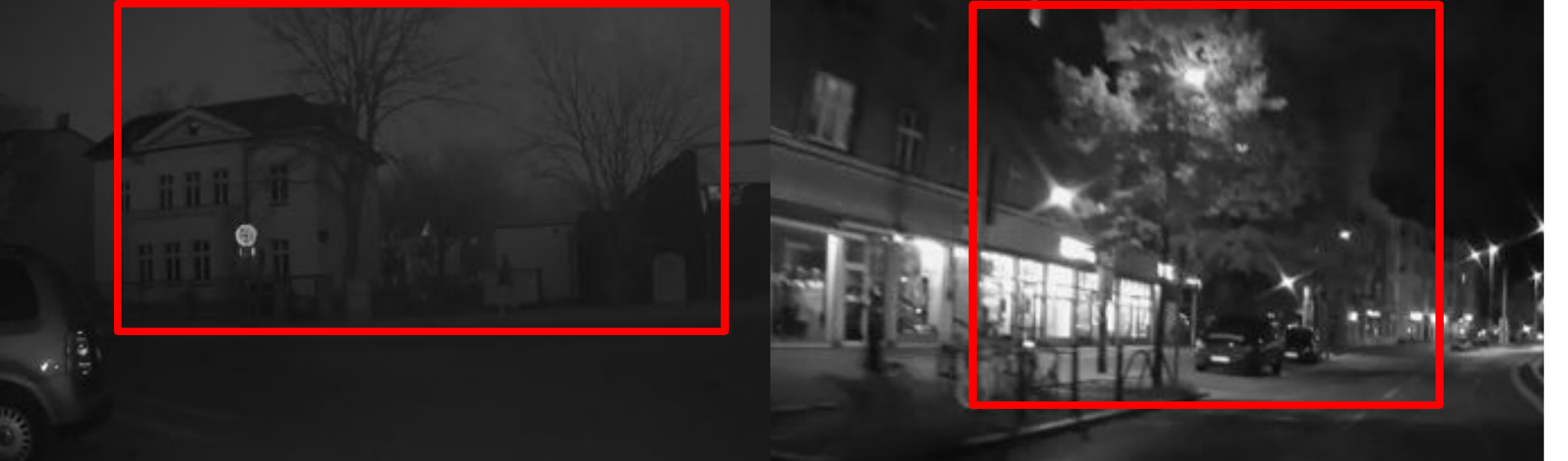}
\end{center}
   \caption{Dynamic lighting effects. The sky is lighter than the surrounding trees if there is no major light source in the scene. Trees are lighter if there are light sources like street lamps in the scene}
\label{fig:dynamiclighting}
\end{figure}

\section{Similarities and Difficulty Level}
Most of the enhancement tasks require multiple transformation subtasks. We consider three distinctive subtasks. Colour mapping/reconstruction, structural reconstruction and noise reduction.
The applications covered in this work require all of the aforementioned subtasks but in different amounts. An overview is shown in Table~\ref{tbl:tfsubtasks}.

\begin{table}[h]
   \footnotesize
    \centering
    \caption{Difficulty of the transformation subtasks for each application. More plusses indicates a higher difficulty level}
    \label{tbl:tfsubtasks}
	\begin{tabular}{lccc}
        	&	 Noise  & Structural  & Colour mapping/\\
        Application		& reduction & reconstruction & reconstruction\\
        \hline
        Removing fog  &	+  &++	&+	\\
        Removing rain  &	+++  &++	&+	\\
        Night to day &	+  &+++	&+++
    \end{tabular}
\end{table}

Images containing fog require the aforementioned transformations but in a moderate amount. Usually, details in the foreground of the image are clearly visible whereas objects at a further distance lose their colours and structural details in proportion to their distance to the camera. 
In case of rainy scenes, the biggest problem is the noise made by rain streaks and the deformations made by raindrops. Colours are slightly greyer than in a clear scene.

The night-to-day transformation is the hardest problem. 
In the black areas the structure is completely lost or the weak features are drowned in noise.
A major problem with night to day vision is the dynamic lighting effects caused by the streetlights and car lights, as is illustrated in Figure~\ref{fig:dynamiclighting}.  In very dark scenes, without a lot of street lights, the sky is typically lighter than the surrounding objects. With street lights and car headlights, the sky is typically black. The same holds for tree tops and bushes. Light reflection can cause part of a tree top to lighten up and without street lights it is typically darker than the sky. This makes the colour mapping extremely difficult.

\section{Transfigurative and Mapping Artifacts}
In this section we describe two common types of artefacts that occur when the vanilla CycleGAN method is used and the main causes of the artefacts. 

In the basic unsupervised adversarial learning method for domain transfer two transformers and two discriminators are trained simultaneously. 
If we train our domain transfer system with a set of unpaired images of daylight and night traffic scenes in a traditional approach, then the goal of the A2B generator is to transform the night image into a day scene and the B discriminator is tasked with making a distinction between real and generated day images. However, it is not explicitly defined that it should keep the original structure in the night scenery. So in practice, the images produced by the generator are completely transfigured. If the system is trained while also trying to minimising the cycle consistency loss~\cite{cyclegan}, then the results improve. Transforming an image from domain A to domain B and transforming it back to domain A should produce an image close to the original image. The A2B transformation should be the inverse of the B2A transformation and vice versa. Some of the structure is implicitly kept because of the cycle consistency loss, but there are still severe changes in the semantics of the scene which makes it practically unusable. The night to day transformation deforms cars and pedestrians as can be seen in Figure~\ref{fig:qualcomp_original_edge}, ~\ref{fig:n2d_p1}, ~\ref{fig:n2d_p2}, ~\ref{fig:n2d_p3} and ~\ref{fig:n2d_p4}. The cycle consistency loss does not guarantee that the semantic structure of the input image will be similar in the different domains. It only imposes that the A2B generator and B2A generator transformations should be inverses of each other

Another frequently occurring problem is incorrect mappings.
This is a major problem for night to day transformation. An important cause is the dynamic lighting effects caused by the streetlights and car lights, as is illustrated in Figure~\ref{fig:dynamiclighting}.  In very dark scenes, without a lot of street lights, the sky is typically lighter than the surrounding objects. With street lights and car headlights, the sky is typically black. The same holds for tree tops and bushes. Light reflection can cause part of a tree top to lighten up and without street lights it is typically darker than the sky. 
These problems occur more prominently in domains that are further apart from each other, because an easy colour transformation is not possible. 
The incorrect mappings are not really discouraged by the CycleGAN loss scheme.

\section{Minimising the Perceptual Distance} 
\label{sec:perc}
The first method to retain as much of the original image meaning during domain transfer is minimising the perceptual distance between the generated image and the original image.
The perceptual distance is calculated by feeding both the original and transformed image through a pre-trained classification network. The weights of the classification network are fixed. Feature maps at different spatial resolutions are taken and compared. A predefined distance metric is used to compare these feature maps between original and generated image. The final perceptual distance is a weighted sum of the distances between the feature maps at different spatial resolutions. By minimising the perceptual distance we implicitly force the transformed image to contain the same meaning as the original image as perceived by the pre-trained classification network.

We are not the first to introduce perceptual distance which has also been called perceptual loss or content loss in previous work. However it has mainly been used in supervised problems. Example applications are increasing the resolution with SRGAN~\cite{superresolution} and transferring style from one image to another. A form of perceptual distance was first used in~\cite{texturegeneration}. The authors used layers of pre-trained classification network to generate textures. In~\cite{bousmalis2016unsupervised} Bousmalis et al. use a masked pairwise mean squared error in an unsupervised domain transfer application similar to our approach but with a different metric.

\begin{figure}[t]
\begin{center}
  \includegraphics[width=\linewidth, trim={0cm 5.9cm 0cm 0cm}, clip]
    {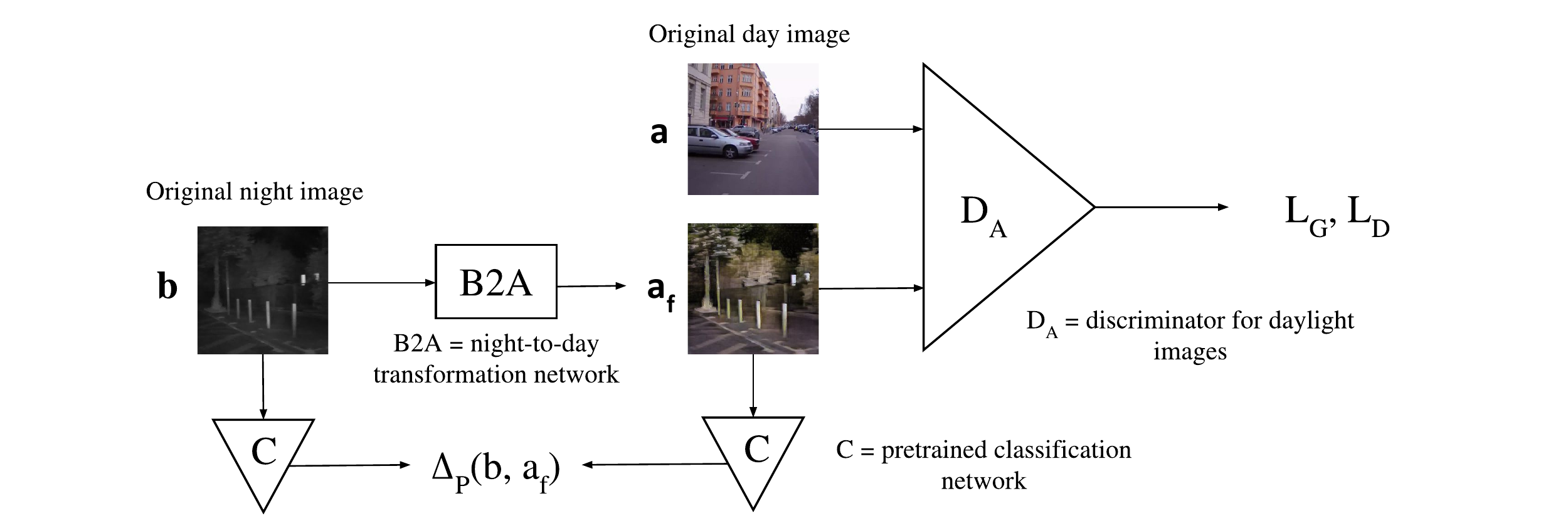}
\end{center}
   \caption{The perceptual distance is calculated by feeding the original image and the transformed image through a pretrained classifier (C) and calculating the distance between the intermediary feature maps.}
\label{fig:perception_loss_schema}
\end{figure}

In Figure~\ref{fig:perception_loss_schema}, a schema summarises how the perceptual distance is minimised during domain transfer.
The original and transformed image are fed to a pre-trained 19-layer VGGNet. 
In previous work, the activations of the convolutional layers right before the maxpool layers are chosen.
For each feature map the mean squared distance is calculated between the activations caused by feeding the original and transformed image. 
The total perceptual distance is then a weighted sum of the L2 distances.
This perceptual distance is used as a loss function and added to the transformation loss.

\begin{align*} 
L_G^{A2B} &= L_{G}(D_B(G_{A2B}(a))) + \lambda_{feat}^A \cdot \Delta_{P}(a, b_f)
\end{align*}
with $\Delta_{P}(a, b_f)$, the perceptual distance between the original image from domain A and the generated image.

This works well for images that are degraded by noise. An example is removing rain drops and streaks. Depending on the size of the noise features we can adjust the weights. If there are only very small noise features we only have to reduce the weight of the first layers. A downside of this approach is that the weights are new hyper parameters. It is cumbersome to optimise and trade off these weights. The higher abstract features are important to keep the high level semantics intact, but lower hierarchy features are important for spatial precision. To overcome this we replaced the VGGNet with a residual network. We only used the activations of the last convolutional layer in a pre-trained 34-layer residual network to determine the perceptual distance. 
In residual networks smaller features better propagate to higher levels because of the skip connections, so there is already a better implicit weighing between smaller and larger features. The Resnet features performed better than our manually fine tuned VGG-based perceptual distance functions.

\begin{figure}[t]
\begin{center}
  \includegraphics[width=\linewidth, trim={0cm 9cm 3cm 0}, clip]
    {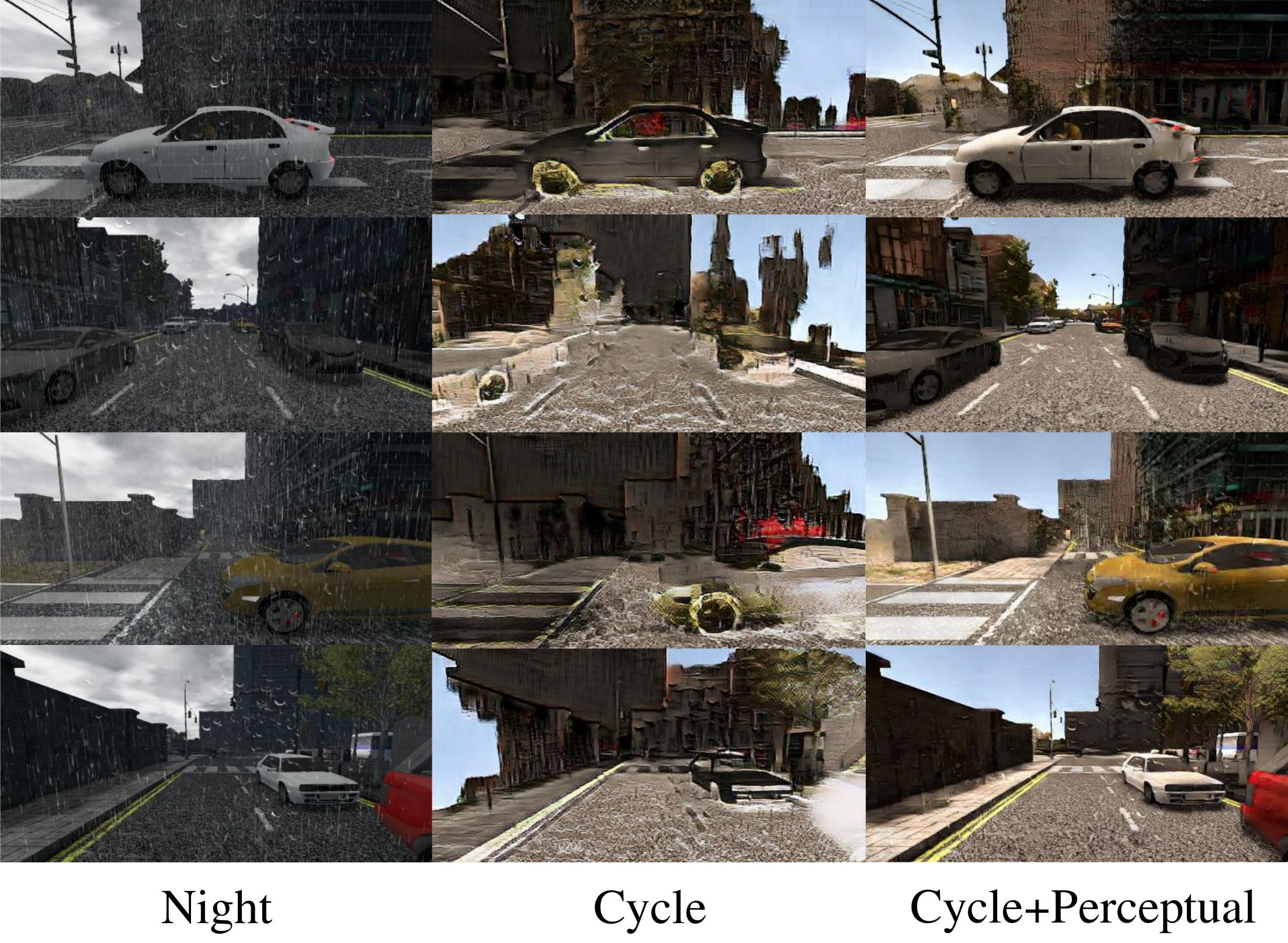}
\end{center}
   \caption{Rain-to-clear with and without minimising the perceptual distance}
\label{fig:r2s_ploss}
\end{figure}

In Figure~\ref{fig:r2s_ploss} a comparison shows the clear difference between the results of a rain-to-clear transformation network trained with and without minimising perceptual distance. 

Unfortunately, minimising the perceptual loss only works if the domains are close to each other in terms of structure and if both domains are close to the domain for which the perceptual model is trained. It fails often for the night-to-day application, because the classifier is not trained with near-infrared images of night scenes.

\begin{figure}[t]
\begin{center}
  \includegraphics[width=\linewidth, trim={0 7cm 0.3cm 0}, clip]
    {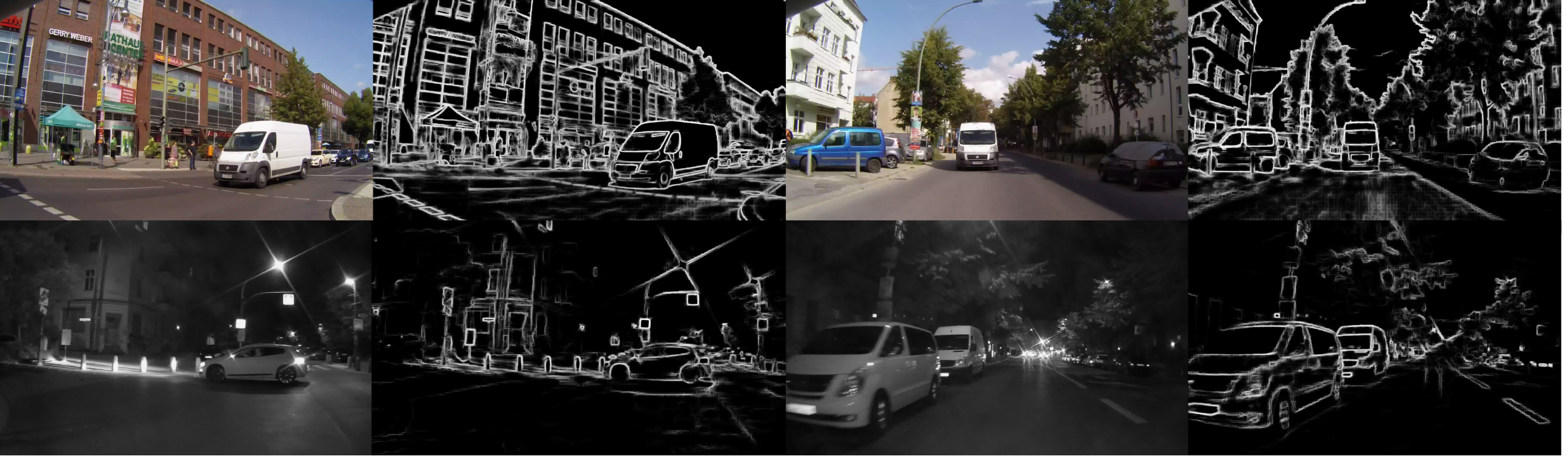}
\end{center}
   \caption{Examples of edge maps produced by our edge detection model}
\label{fig:edgemaps}
\end{figure}

\section{Edge Features}\label{sec:edge}
A large part of the semantics of an image can be summarised by its edge features. A human can easily distill the content of an image by looking at an edge map of the image, an example is shown in Figure~\ref{fig:edgemaps}. In some domain-transfer applications it is desirable that the edge features are preserved. An example is the night to day transformation. A counter example is brightening and clarifying rainy scenes. For this application, it is not desirable to keep edges during transformation, because rain streaks have strong edge features.

As can be seen in Figure~\ref{fig:edgemaps}, the edge map of the daylight scene and the  night scene are hard to distinguish. It is difficult to distinguish whether it is based on a daylight or night scene. The one clear difference is that daylight photos typically produce more edges compared to night photos. So the edge features can be used to guide the learning process for the domain transfer models. We propose the following edge feature scheme for the night-to-day. The night-to-day transformer should preserve the edges of the night image. This guides the transformer to produce daylight images with the same structure. For the day-to-night we propose to force the transformer to not introduce new edges, which prevents it from hallucinating new structures and thereby changing the semantics of the scene. For the removing fog application we propose to use the same scheme. The fog-to-clear transformation should preserve even the slightest hint of an edge during the fog-to-clear transformation, but there should not be any new edges introduced by the clear-to-fog transformation.

Several published articles described on a convolutional network architecture for detecting edges in an image. Notable examples are PixelNet~\cite{pixelnet} and Holistically-Nested Edge Detection (HED)~\cite{hed}. We tried both and choose the latter to base our edge detection model on. The reason being, HED produced edge maps with more fine edges and details than PixelNet. This is also indicated by Bansal et al. in~\cite{pixelnet}, where both edge detection models are compared. The original HED architecture uses the pre-trained feature layers of a 16-layer VGG network. A VGGNet is relatively memory intensive, so we replaced it with the feature layers of a residual network with 34 layers.

\begin{figure}[t]
\begin{center}
  \includegraphics[width=\linewidth, trim={2cm 3.2cm 2.5cm 0cm}, clip]
    {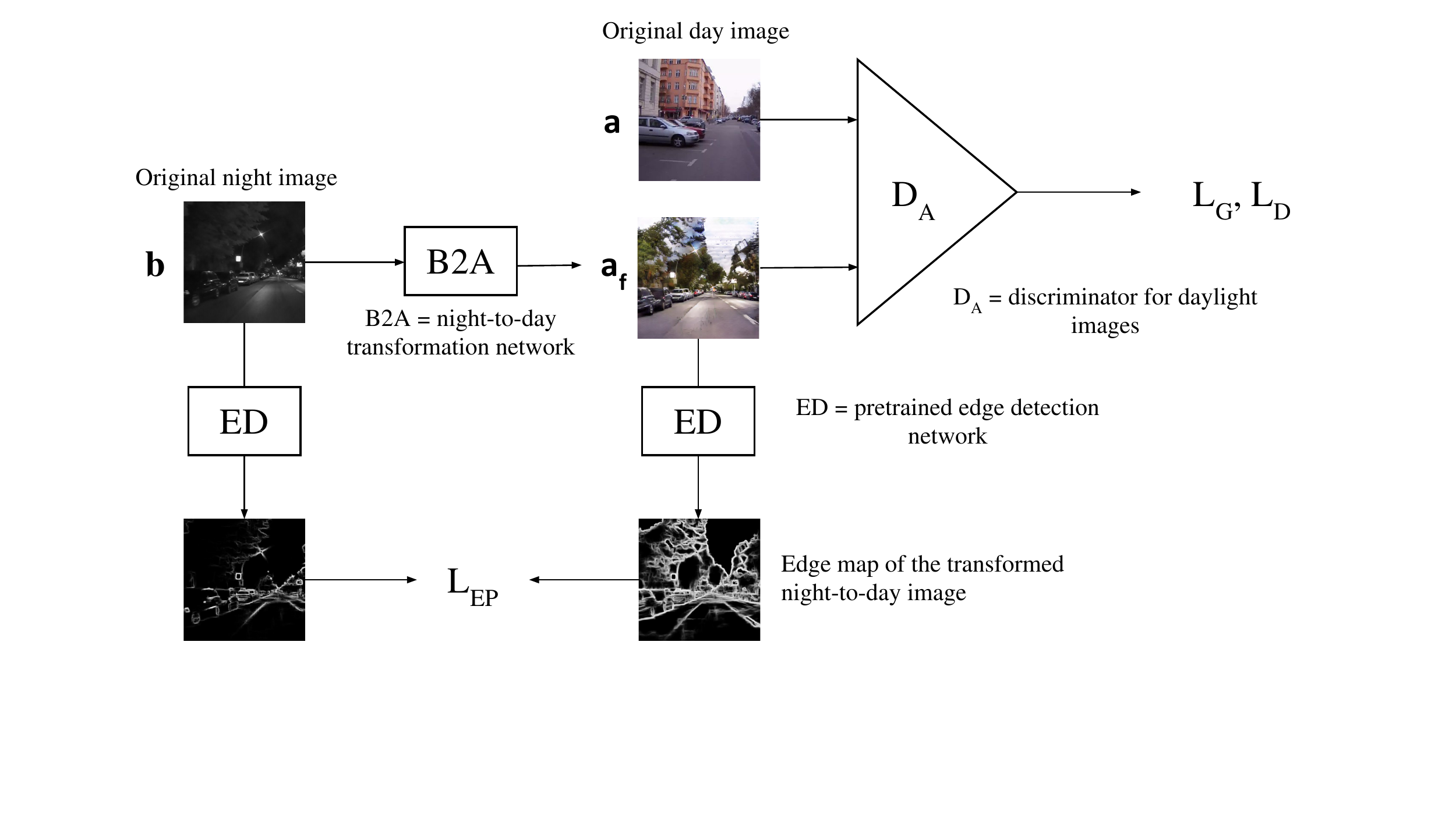}
\end{center}
   \caption{The edge maps of the original and generated image are produced by the edge detection model (ED). The edge maps are used to calculate the edge preservation loss }
\label{fig:edge_loss_schema}
\end{figure}

Both the original image ($b$) and the generated image ($a_f$) are fed to the edge detection model, as shown in Figure~\ref{fig:edge_loss_schema}. The resulting edge maps are then used to define the following edge feature based losses. They are based on the squared error between both edge maps, because small changes in edge features are relatively less important than major changes.
The first one is the edge preservation loss ($L_{EP}$).
\begin{gather}
\text{err}(b,a_f) = \text{ED}(b)-\text{ED}(a_f) \\
\text{pos}(b,a_f) = \frac{1+\text{sign}(\text{err}(b,a_f))}{2} \\
f_{\text{bal}} = \frac{\sum_{i,j}(1-\text{ED}(b)(i,j))}{W\cdot H} \\
L_{EP}(b,a_f) =  f_{\text{bal}} \cdot \norm{\text{pos}(b,a_f) \cdot \text{err}(b,a_f)}_{2}
\end{gather}
with ED, the edge detection model, W and H, the width and the height of the image.

The second factor indicates that only if there is a positive error, the loss should be taken into account. Only if the edge features are less strong in the transformed image the loss is larger than zero. The transformer should not be punished if the edges are stronger in the target domain. Any hint of an edge could be important for the reconstruction of the scene, and typically they should be made stronger in the night-to-day and fog-to-clear transformation. The first factor is a balancing factor. If there are only a small amount of edges in the edge map, then it is more important that they are preserved when there are a lot of edges in the edge map. 

The edge introduction loss ($L_{EI}$) is the second edge feature based loss and it should prevent the introduction of new edges after transformation.

\begin{gather}
\text{neg}(a,b_f) = \frac{1-\text{sign}(\text{err}(a,b_f))}{2} \\
L_{EI} = \frac{\sum_{i,j}\text{ED}(x)(i,j)}{W\cdot H} \cdot \norm{neg(a,b_f) \cdot err(a,b_f)}_{2}
\end{gather}

In the second factor the sign function is used to only account the squared error between the edge maps when there is a negative error. This means that this error will only be taken into account when the edges are stronger in the transformed image than in the original image.
The first factor is the balancing factor. It is based on the number of edges in the input image. Edges are relatively sparse features. 



\begin{figure}[t]
\begin{center}
  \includegraphics[width=\linewidth, trim={2cm 6.5cm 1.6cm 0cm}, clip]
    {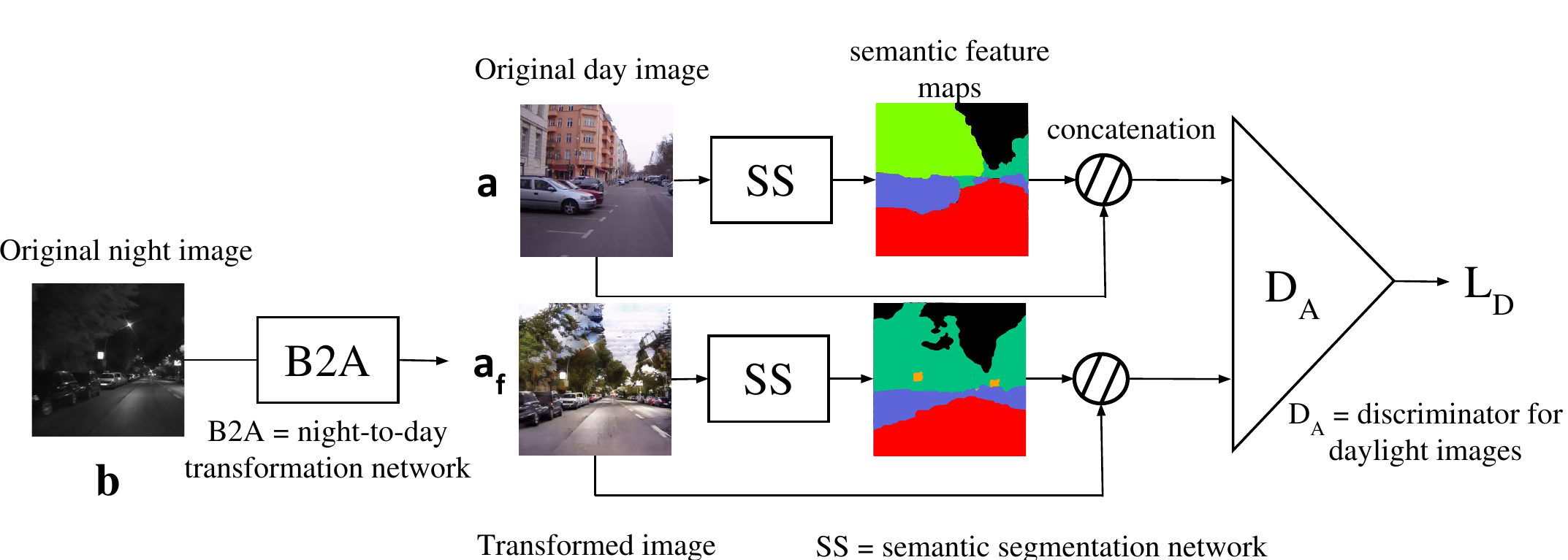}
\end{center}
   \caption{Semantic segmentation maps are calculated for both the real image and the transformed image and they are concatenated and fed to the discriminator  }
\label{fig:ss_schema}
\end{figure}

\section{Generating Semantically Sound Images}
\label{sec:ss}
Although transferring the edge features majorly improves the resulting transformed images, there are still some semantic mistakes being produced by the transformer. Common artefacts include buildings with trees built in and trees with some building features in between the branches without there being a building behind the tree. An example can be seen in Figure~\ref{fig:qualcomp_edge_ss}.

To generate images with fewer semantic mistakes, we suggest to exploit expert models of one of the domains, typically the clean domain. We propose to provide more relevant information to the discriminator to implicitly force the transformer to produce images that are semantically correct or at least more plausible.

An ideal candidate for an expert network is a semantic segmentation network. Additionally the segmentation maps have the same resolution as the input image, so they can easily be concatenated with the input image before feeding it to the discriminator.
Other expert networks like classifiers could also provide more semantic information to the discriminator. Intermediary feature maps could be fed to the discriminator, but this requires changes to architecture because the feature maps typically have a smaller spatial resolution than the input image.

In Figure~\ref{fig:ss_schema} there is an illustration of the proposal. If we have a pre-trained segmentation model for the clear domain, we can use it to generate segmentation maps for both the original images in the clear domain and the generated fake images. We feed images and their respective segmentation maps to the domain discriminator. The domain discriminator is then trained to distinguish between real and generated images. If the segmentation map of the generated image does not make sense, it will be easier for the discriminator to detect the fake images. The result is an image with fewer semantic mistakes.
An additional advantage of feeding the segmentation maps to the discriminator is that they stabilise the training process. The segmentation maps are high quality features that are not trainable. 

Several publications focus on segmenting traffic scene imagery.
Approaches to semantic segmentation have been published in \cite{tiramisu}, \cite{fcn} and \cite{unet}.
We chose a FC-DenseNet architecture, because it is a recent good performing segmentation architecture and requires few parameters. We trained it to segment the traffic imagery in the CityScapes dataset~\cite{cityscapes}. 
One small architectural change helps the discriminator even more. A segmentation network typically outputs a binary segmentation map per class.
The last nonlinearities of a FC-DenseNet are pixel-wise softmax. We changed this to sigmoid nonlinearities. It enables the model to output whether a pixel is part of both a building and a tree, which is not possible in real images, but it can occur for generated images and this helps the discriminator to find inconsistencies.

\section{Improvements to the Architecture}
\label{sec:arch}

There are two common transformation network architectures for Cycle GAN. A U-Net based architecture~\cite{unet}, an architecture originally used for segmentation tasks, and an architecture built with residual blocks~\cite{resnet}. We propose to use an architecture built based on the FC-DenseNet model~\cite{tiramisu}, which is also used for segmentation. It outperforms U-Net based segmentation networks and is based on the dense block architecture.

We tweaked the architecture to make it more suitable for transforming images. In the original FC-DenseNet architecture, the spatial resolution is increased by transition up modules. The transition up module consists of a 3x3 transposed convolution layer. The upsampled feature maps are concatenated with the feature maps coming from the skip connection. Together they form the input to a new dense block.
We replaced the transposed convolution upsample method with sub-pixel convolutional upsampling, which is first introduced in~\cite{subpixel} for increasing the resolution of video and images.

For the night to day application, we noticed some artefacts where a large part of the scene is black. The transformer defaults to the most probable mapping for black areas, which is typically the sky or tree top patches. If this black patch does not occur at the top of the image this can lead to unrealistic transformations. To overcome this issue the field of view is maximally enlarged. Adding more convolutional layers increases the field of view linearly, while introducing more parameters and a higher memory consumption. Adding more downsampling levels is more effective, because it quadratically increases the field of view. However, it is limited by the size of the input patch. We increase the number of levels to match the size of the input patch and keep the same number of layers per level. The number of downsampling levels and the training crop sizes used during training is reported in Table~\ref{tbl:scales}.

\section{Experiments}
\subsection{Datasets}
To create a labeled dataset we used a simulation environment based on GTA V v1.4 enhanced by several mods to increase photorealism and weather simulations. 
We further modified GTA to be controllable by an external script.
It allowed us to automatically collect images from many different sceneries and weather conditions. Without a simulation, it would not be feasible to gather image pairs of different weather conditions. Even though all experiments are performed under unsupervised conditions, having image pairs in the validation set allows to better assess the performance of the proposed techniques. For the night scenes, we developed a shader for an infrared night vision effect, based on~\cite{nightvisionshader}. For fog we used the volumetric effect provided by the NaturalVision mod.
The rainy weather condition in modified GTA engine did not provide a convincing effect, so we used part of the public dataset called Synthia1~\cite{synthia}. Synthia1 contains images of traffic scenes. Some of the images are taken during rain showers. The images have strong rain streaks including raindrops on the camera lens. After selection and preprocessing, the dataset contains roughly 500 rainy and 500 sunny training images. The test set contains 75 images each. The downside is that the dataset does not have corresponding pairs.

We also collected a dataset with real world night images. 
Images were captured while driving around in an urban environment.
During the day a regular RGB camera and during the night a Near Infra Red camera was used.
After curing the dataset contains around 5000 night images and 5000 day images.

\subsection{Quantitative Comparison}
The proposed techniques are benchmarked with the labeled datasets.
We compare four different configurations.
\begin{itemize}
\item \textbf{Cycle+Resnet}: the baseline configuration with a basic cycle-consistency adversarial approach with a generator built out of residual blocks
\item \textbf{Edge+Resnet}: In this configuration the edge losses are added during training
\item \textbf{Edge+FCDenseNet}: The edge losses and our new FCDenseNet based transformation architecture described in the previous section.
\item \textbf{Perc+FCDenseNet}: The new FCDenseNet based transformation architecture trained while minimising the perceptual distance
\end{itemize} 
For the exact parameters of the configurations we refer to the additional materials section.

\begin{figure}[t]
\begin{center}
  \includegraphics[width=0.9\linewidth, trim={0 0.25cm 0cm 0}, clip]
    {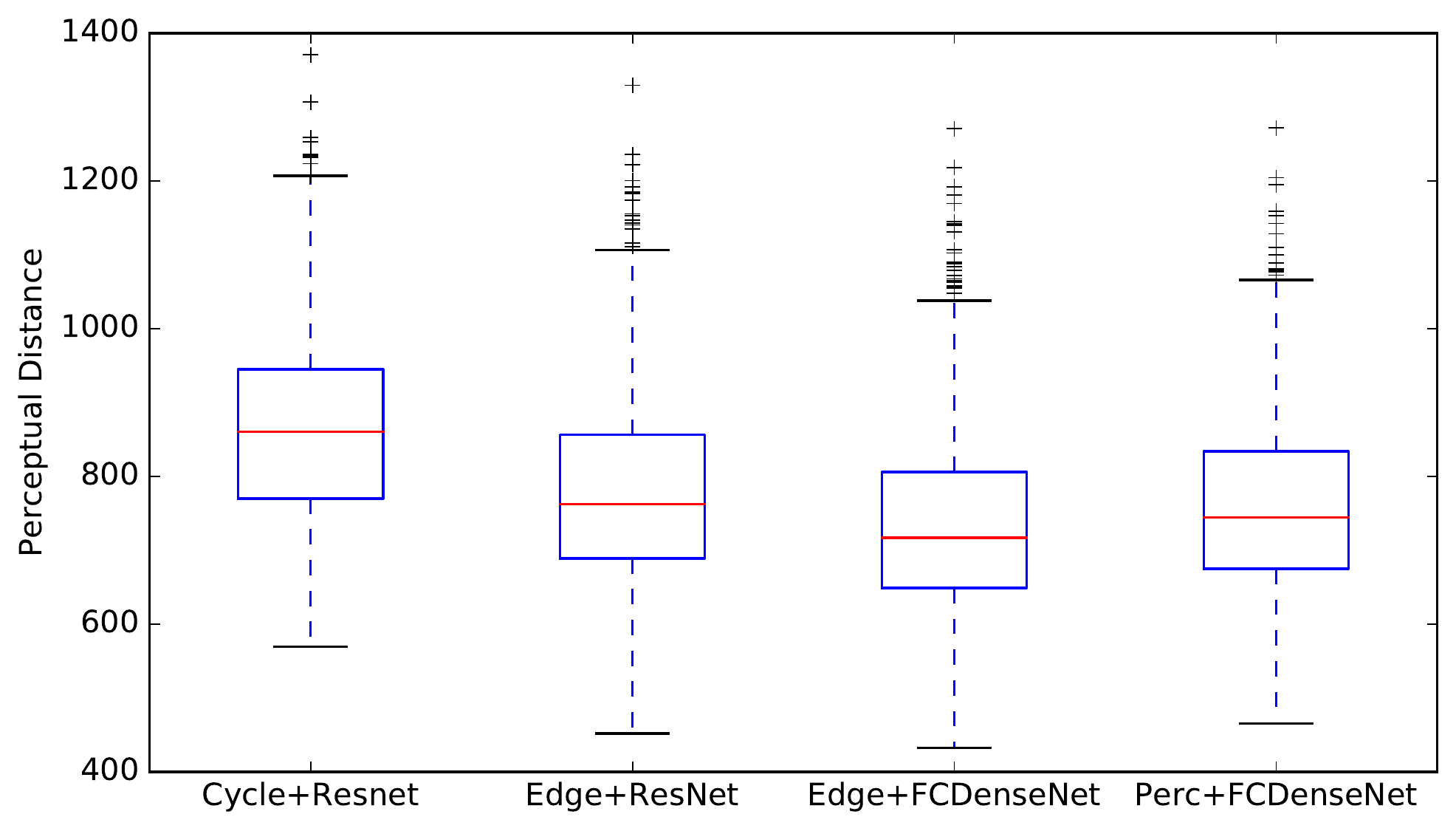}
\end{center}
   \caption{Results for the removing fog task} 
\label{fig:boxplot_fog}
\end{figure}

\begin{figure}[t]
\begin{center}
  \includegraphics[width=0.9\linewidth, trim={0 0.25cm 0cm 0}, clip]
    {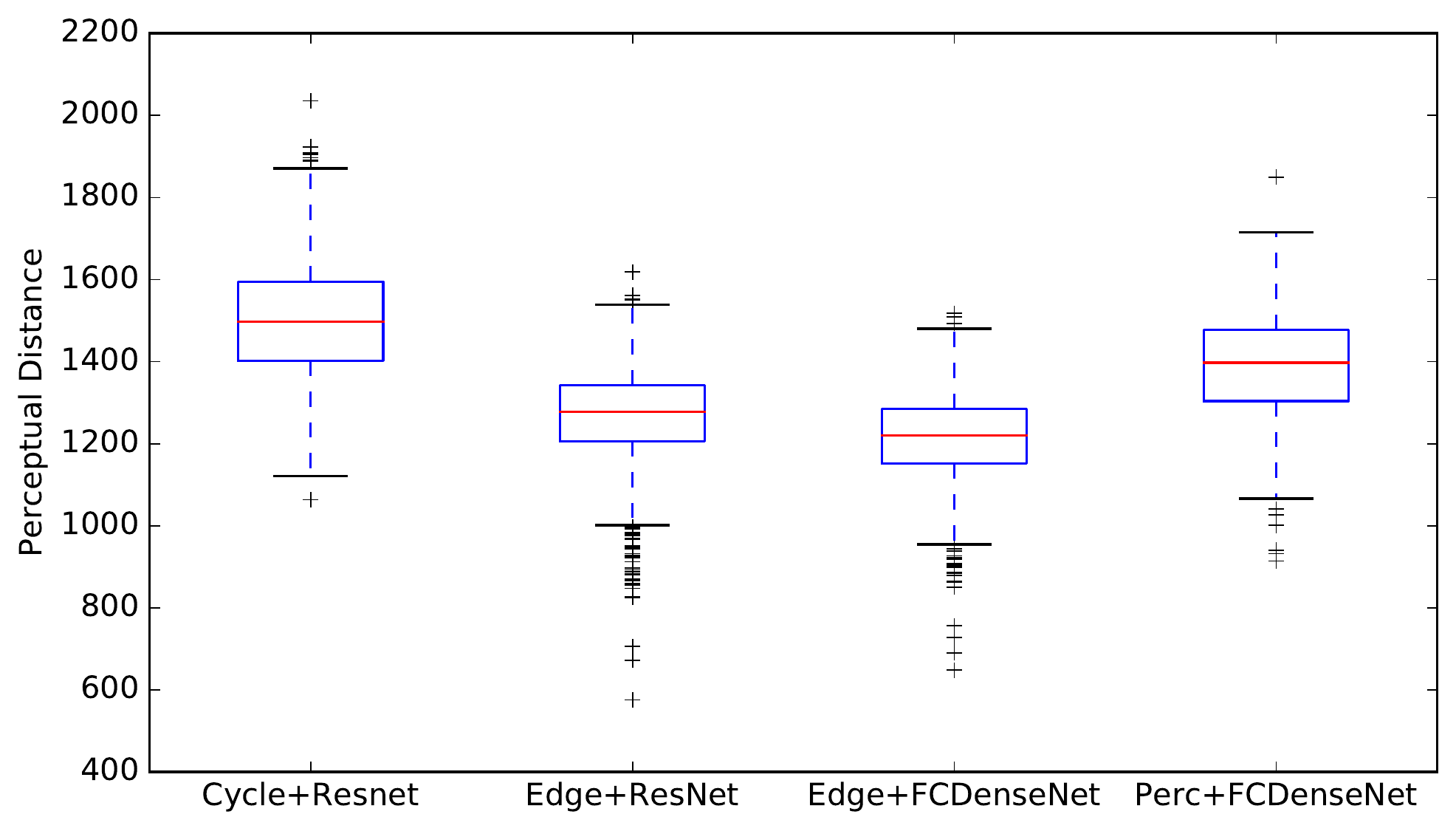}
\end{center}
   \caption{Results for the night to day transformation task} 
\label{fig:boxplot_ir2day___}
\end{figure}
To benchmark the configurations the perceptual distance between the transformed image and the real image is calculated. We used a pre-trained 16-layer VGGNet and used equal weights for the different feature maps.

In Figure~\ref{fig:boxplot_fog} and Figure~\ref{fig:boxplot_ir2day___} the box plots summarise the results of the different configurations. 
For both applications, our proposed techniques clearly outperform the baseline and the \textit{Edge+FCDenseNet} configuration is the best performing configuration.
Another similarity is that the FCDenseNet architecture also outperforms the ResNet based architecture.
The biggest difference between the both tasks is that minimising the perceptual distance works almost as good as the \textit{Edge+FCDenseNet} configuration for removing fog, but clearly does not help much in case of the night-to-day transformation.

We also observed that there is a large difference in average perceptual distance for the two applications. The average distance for the night-to-day transformation is higher than the fog-to-clear transformation. This is an indication of how difficult the night-to-day transformation is compared to the fog-to-clear transformation.

\subsection{Qualitative Comparison}
\begin{figure}[t]
\begin{center}
  \includegraphics[width=\linewidth, trim={0 108cm 1.5cm 0}, clip]
    {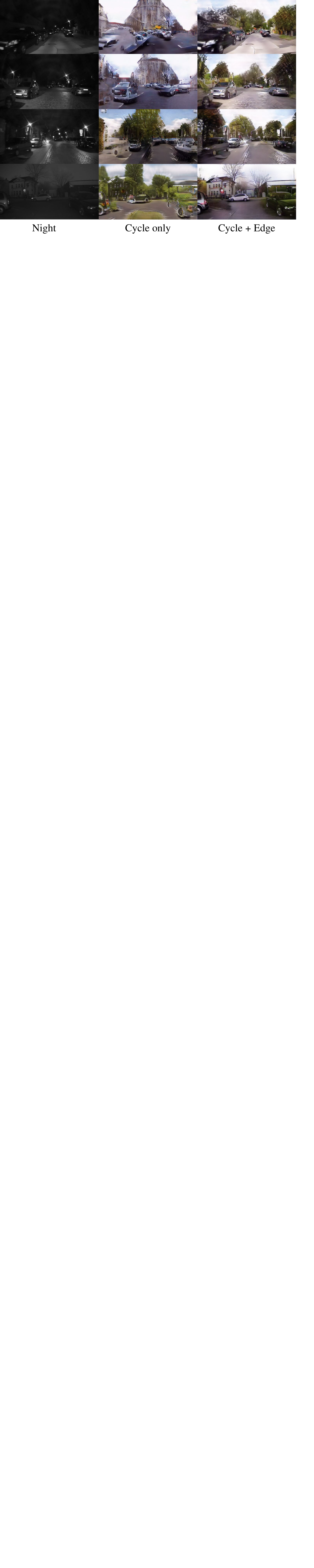}
\end{center}
   \caption{Qualitative comparison between CycleGANs and our approach with edge feature losses} 
\label{fig:qualcomp_original_edge}
\end{figure}

In the qualitative comparison we compare the \textit{Edge+FCDenseNet} configuration with the original \textit{Cycle+Resnet} configuration on the real image dataset for the night-to-day transformation task. A few samples are shown in Figure~\ref{fig:qualcomp_original_edge}. For more high resolution samples we refer to the additional materials section.
The transformation results of the \textit{Cycle+Resnet} approach typically have transmutation artefacts. The model hallucinates new buildings and changes the perspective in the image. Cars are deformed and the semantics of the scene are changed. Our new technique clearly outperforms \textit{Cycle+Resnet}. The structure is not changed. The scene is overall brighter and it is easier to understand the semantics of the scene. It still has some problems with the glares coming from street lights. They produce sharp edges, as can be seen in Figure~\ref{fig:edgemaps}, and these edges are enforced in our training process. This leads to unnatural transitions between air and tree tops. Another problem is the completely dark areas where our model sometimes hallucinates dark green tree foliage. 

\begin{figure}[t]
\begin{center}
  \includegraphics[width=\linewidth, trim={0 9cm 2.5cm 0}, clip]
    {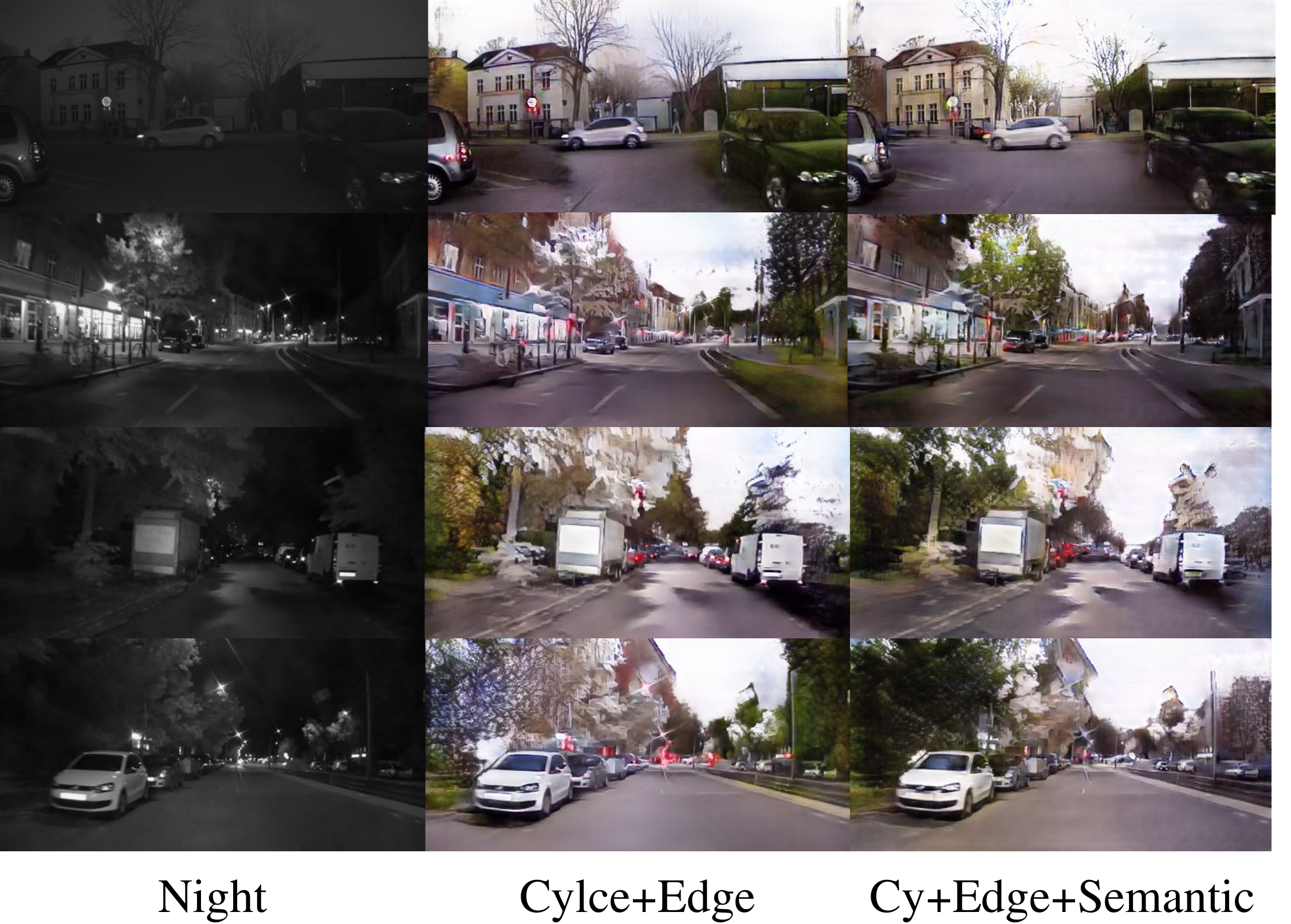}
\end{center}
   \caption{Qualitative comparison between our approaches with edge feature losses and with/without semantic segmentation maps fed to the discriminator. The first two examples show improvement, the two last examples do not show a remarkable improvement} 
\label{fig:qualcomp_edge_ss}
\end{figure}

To further improve the semantics of the produced images we proposed to feed semantic segmentation maps to the discriminator.
In Figure~\ref{fig:qualcomp_edge_ss} some samples are shown. We compare the \textit{Edge+FCDenseNet} configuration with and without the semantic maps feed to the discriminator.
For some samples this helped the colour precision as can be seen in the top two rows in Figure~\ref{fig:qualcomp_edge_ss}. If we further zoom in on the images in the second row, then you can observe trees illuminated by street lights have a white colour with the old configuration and in the new configuration this changed to green. The train tracks  that were coloured green are now correctly coloured grey. However, for half of the images the semantic feed did not really improve the result as can be seen in the bottom row images.

\section{Conclusion}
We introduced new techniques to improve the training of domain transfer models, targeted to enhance images of traffic scenes taken in conditions with limited visibility. We covered degraded visibility caused by rain, fog and the lack of light during the night.

For domains that are close enough in terms of colouring, we propose to force the generator to minimise the perceptual distance between the original image and the transformed image. This is applicable to images of rainy and foggy scenes. 
For domains that are further apart, like daylight and near infrared images during the night, 
we propose two separate techniques that can be used in tandem.
In the first technique we use the edge features of the original image and the transformed image.
It can be beneficial to force the same edge features after transformation or to prevent any new edge features from emerging during transformation.
In the second technique, the discriminator is provided with high quality fixed non trainable features.
A pre-trained semantic segmentation model is an ideal candidate because it gives pixel-wise high quality features that can be easily merged together with the image and fed to discriminator.

We also showed that synthetic datasets generated by gaming engines can help guide us to develop and train a better domain transformation models. We generated images of scenes taken in poor visibility conditions and the same scene in perfect conditions. This allows us to benchmark our unsupervised domain transfer approaches. Finally, we validated the results on a datasets with images of real traffic scenes during the day and the night and showed superior performance compared to the original cycle-consistent adversarial approach.


{\small
\bibliographystyle{ieee}
\bibliography{edge}
}

\newpage
\section{Additional Results and Documentation}\label{sec:add}

\subsection{Parameter Values}
In this section we describe the parameter values used to generate the images produced in this work.
In Table~\ref{tbl:overview_params} an overview is given of the most important parameters and their symbol or abbreviation.  
The parameters that stay constant for all our configurations are summarised in Table~\ref{tbl:constant_params}.
We mainly consider three configurations in this work. For each configuration the parameter values are listed in a separate table. Table~\ref{tbl:cycle} contains the parameter values for the baseline configuration with the cycle consistency loss. Table~\ref{tbl:cycle_pdist} contains the parameter values for the configuration with an additional perceptual distance loss, which corresponds to the \textit{Perc+FCDenseNet} configuration in the experiments section. Table~\ref{tbl:cycle_pdist} lists the parameter values for the configuration with the additional edge preservation and edge introduction losses, which corresponds with the \textit{Edge+FCDenseNet} configuration in the experiments section. The constant parameters and the parameters that have a value of zero are omitted for clarity.

\begin{table}[h!]
   \footnotesize
    \centering
    \caption{Constant parameters }
    \label{tbl:constant_params}
	\begin{tabular}{cc}
        Parameter	& Value\\
        \hline
               
        $N_{iter}$  & 100 \\
        $N_{iter, decay}$  & 100 \\
        
        $Lr$  & 0.0002 \\
        $\beta_{1}$  & 0.5 \\
        
        Pool size  & 50 \\

    \end{tabular}
\end{table}

\begin{table}[h!]
   \footnotesize
    \centering
    \caption{Training parameter values for the $cycle$ approach for the real night to day task.}
    \label{tbl:cycle}
	\begin{tabular}{cc}
        Parameter	& Value\\
        \hline
        $\lambda_{cy, A}$  &	10   \\
        $\lambda_{cy, B}$   & 10   \\

    \end{tabular}
\end{table}

\begin{table}[h!]
   \footnotesize
    \centering
    \caption{Training parameter values for the $cycle+\Delta_{P}$ approach.}
    \label{tbl:cycle_pdist}
	\begin{tabular}{cc}
        Parameter	& Value\\
        \hline
        $\lambda_{cy, A}$  &	10   \\
        $\lambda_{cy, B}$   & 10   \\
        
        $\lambda_{p, afb}$ & 0.25 \\
        $\lambda_{p, bfa}$ & 0.25 \\
        $\lambda_{p, farb}$ & 0.25 \\
        $\lambda_{p, fbra}$ & 0.25 \\

    \end{tabular}
\end{table}

\begin{table}[h!]
   \footnotesize
    \centering
    \caption{Training parameter values for the $cycle+edge$ approach.}
    \label{tbl:cycle_edge}
	\begin{tabular}{cc}
        Parameter	& Value\\
        \hline
        $\lambda_{cy, A}$  &	10   \\
        $\lambda_{cy, B}$   & 5   \\
        
        $\lambda_{ep, afb}$  & 100 \\
        $\lambda_{ep, farb}$  & 100 \\
        
        $\lambda_{ei, bfa}$  & 10 \\
        $\lambda_{ei, fbra}$  & 10 \\

    \end{tabular}
\end{table}

During training the images in the datasets are first downsized and a random crop is taken. 
The datasets with synthetically generated images were preprocessed so that downsizing was not necessary anymore.
The load sizes and the size of the cropped patches for each configuration are summarised in the following table.

\begin{table}[h!]
   \footnotesize
    \centering
    \caption{Resolutions of the images and crop sizes during training.}
    \label{tbl:resolutions}
	\begin{tabular}{ccccc}
	& real/synth. & original & load & crop \\
	\hline
	\hline
	night2day & real & 1920x1080 & 512x288 & 256x256 \\
	night2day & synth & 256x256 & 256x256 & 192x192 \\
	removing fog & synth & 256x256 & 256x256 & 192x192\\
	removing rain & synth & 256x256 & 256x256 & 192x192 \\
    \end{tabular}
\end{table}

\begin{table}[h!]
   \footnotesize
    \centering
    \caption{Crop sizes and number of scales in the FCDenseNet architecture.}
    \label{tbl:scales}
	\begin{tabular}{cccc}
	& real/synth. & crop size & \# scales \\
	\hline
	\hline
	night2day  & real & 256x256 &  8 \\
	night2day & synth   & 192x192 & 6 \\
	removing fog & synth  & 192x192 & 6\\
	removing rain  & synth & 192x192 & 6\\
    \end{tabular}
\end{table}

\subsection{Runtime}
The new approaches, $cycle+\Delta_{P}$ and $cycle+edge$ require more memory and runtime during training. 
Each original and transformed image has to be processed by a pre-trained classifier or edge detection model respectively. 
The extra forward passes through the classification/edge detection model are the cause for the extra runtime.  Additionally the classification/edge detection models are loaded in the GPU so this requires more VRAM. As a consequence the maximal batch sizes decreases, which further impacts the runtime. The total runtimes and maximum batch sizes for the different approaches are reported in Table~\ref{tbl:runtime_memory} for the real night-to-day transformation task. We used \textit{p3.8xlarge} instances from Amazon Web Service. These instances contain 4 NVIDIA Tesla V100 GPUs.
For inference the runtime of the new approaches is very similar to the original $cycle$ method, because we only have to perform a forward pass through the transformation network, which is an advantage if the transformers are deployed in real life applications.

\begin{table}[h!]
   \footnotesize
    \centering
    \caption{Runtime of the train process for the night to day transformation on real data. Training server contains 4 x NVIDIA Tesla V100}
    \label{tbl:runtime_memory}
	\begin{tabular}{ccc}
	Approach & Total train time & Max. batch size \\
	\hline
	\hline
	$cycle$ & 18h21m & 28\\
	$cycle+\Delta_{P}$ & 24h49m & 16 \\
	$cycle+edge$ & 26h00m &16 \\
    \end{tabular}
\end{table}

\subsection{Additional Results}
\begin{figure*}[t]
\centering
  \includegraphics[width=\linewidth, trim={0 0cm 0cm 0cm}, clip]
    {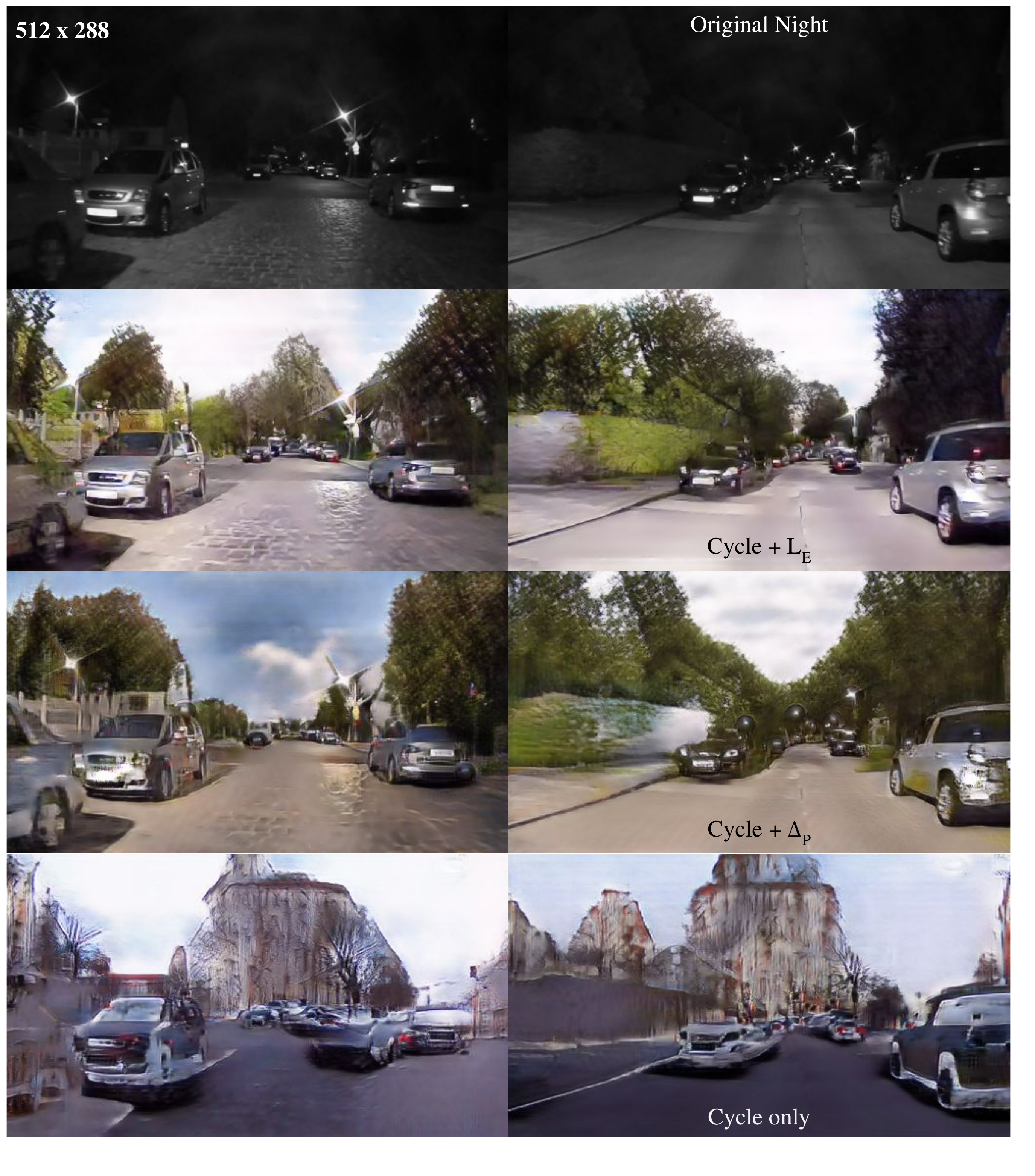}
   \caption{Different approaches for the night to day transformation on real images, part 1/4}
\label{fig:n2d_p1}
\end{figure*}

\begin{figure*}[t]
\centering
  \includegraphics[width=\linewidth, trim={0 0cm 0cm 0cm}, clip]
    {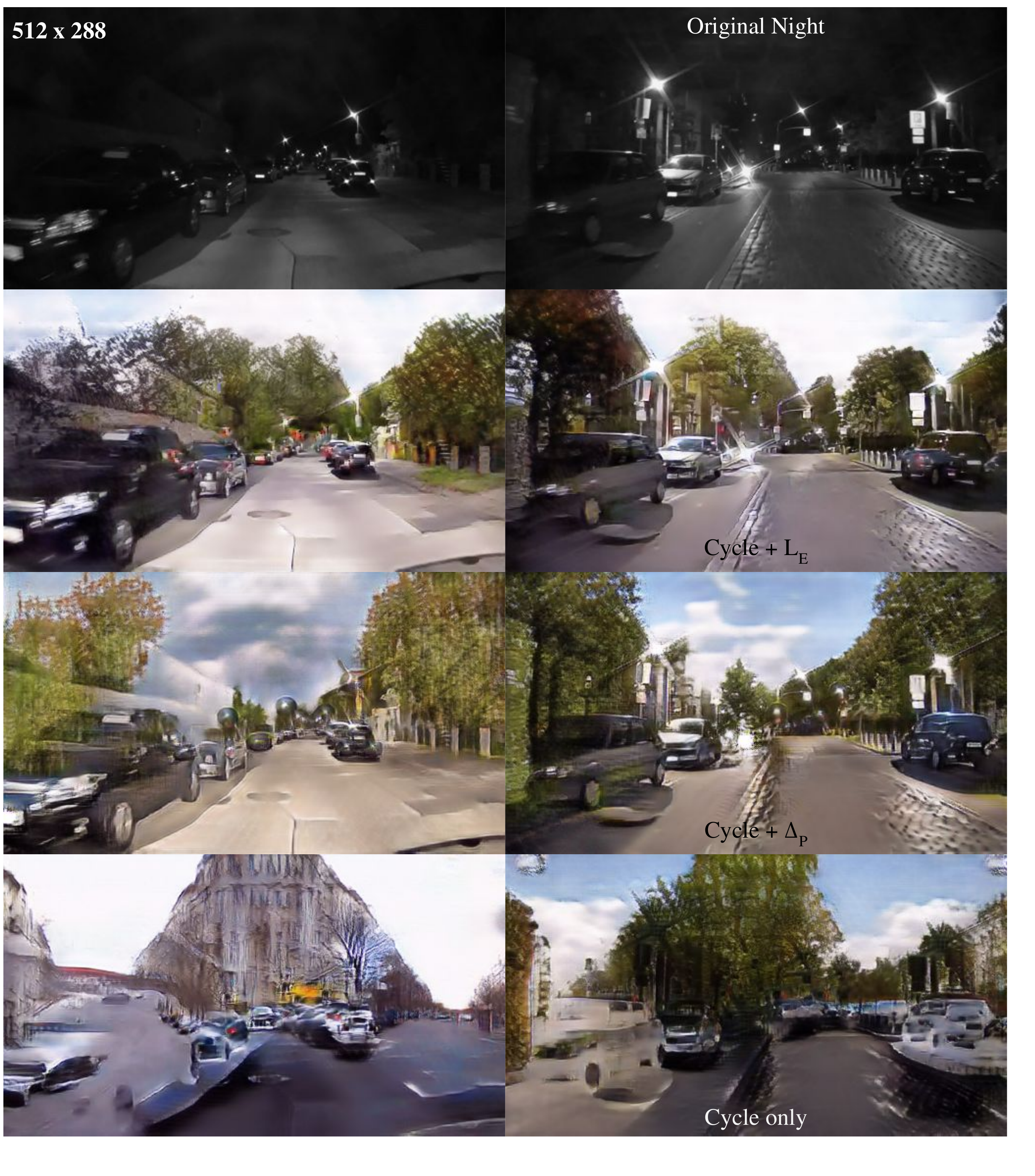}
   \caption{Different approaches for the night to day transformation on real images, part 2/4}
\label{fig:n2d_p2}
\end{figure*}

\begin{figure*}[t]
\centering
  \includegraphics[width=\linewidth, trim={0 0cm 0cm 0cm}, clip]
    {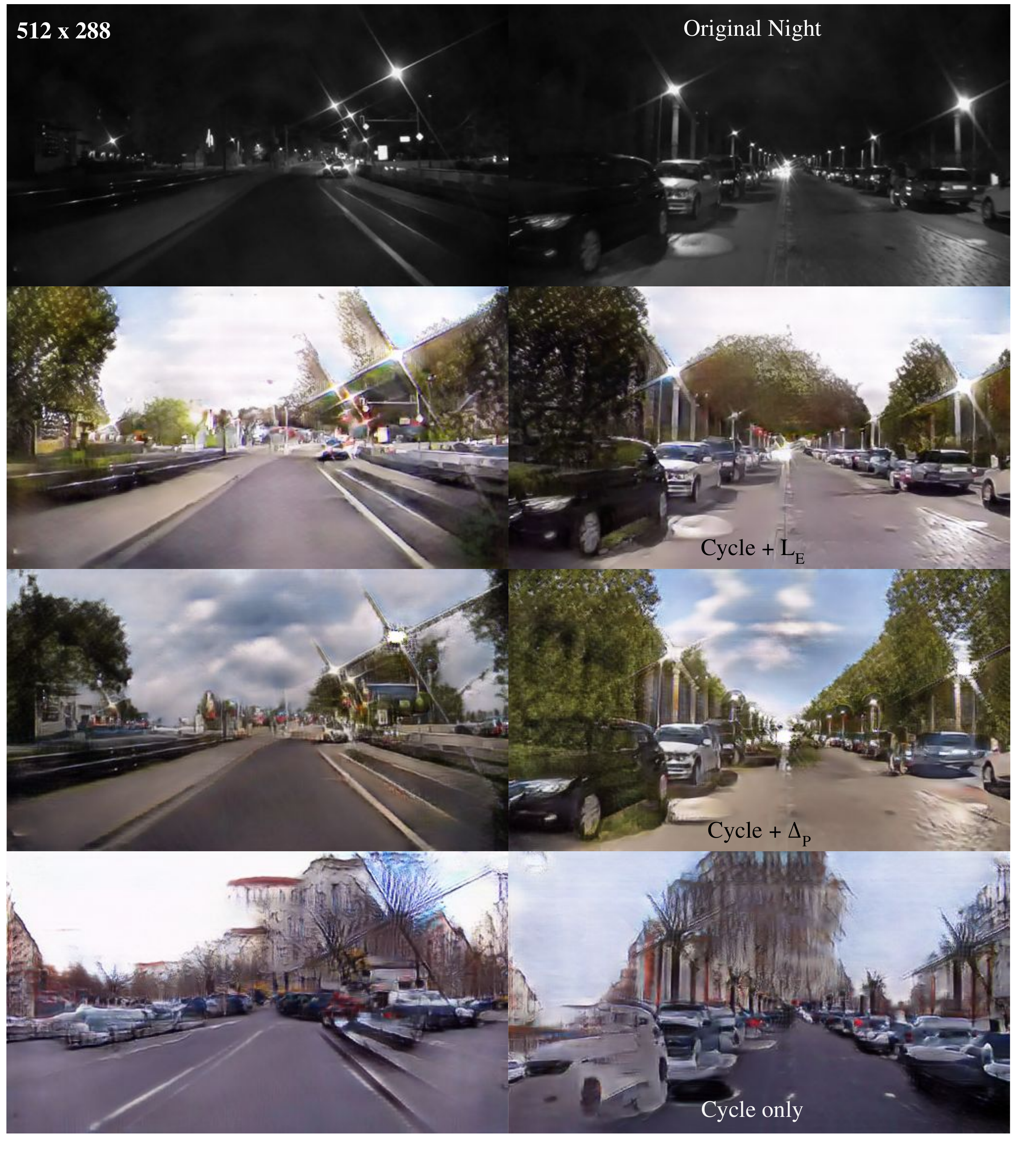}
   \caption{Different approaches for the night to day transformation on real images, part 3/4}
\label{fig:n2d_p3}
\end{figure*}

\begin{figure*}[t]
\centering
  \includegraphics[width=\linewidth, trim={0 0cm 0cm 0cm}, clip]
    {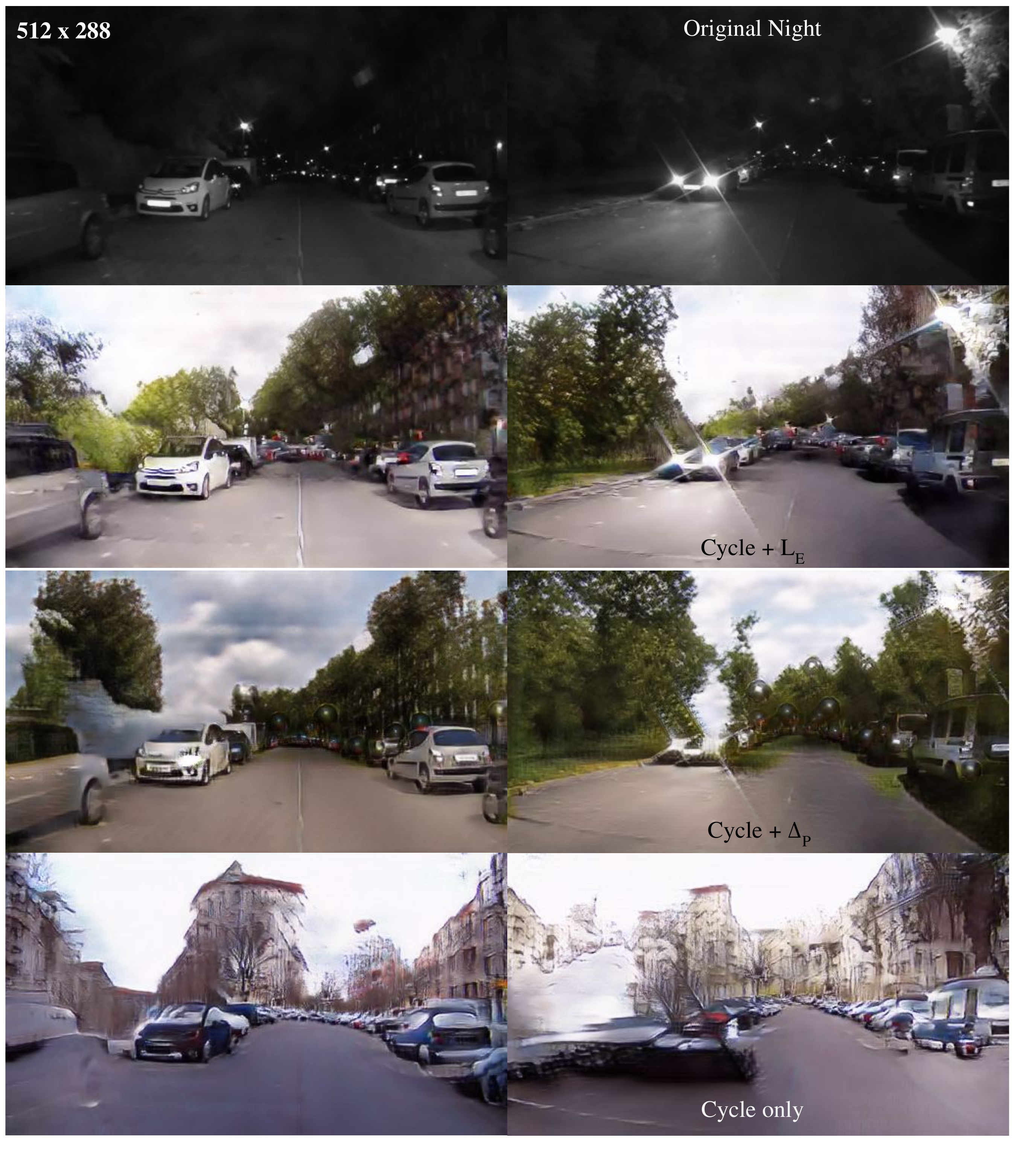}
   \caption{Different approaches for the night to day transformation on real images, part 4/4}
\label{fig:n2d_p4}
\end{figure*}

\begin{figure*}[t]
\centering
  \includegraphics[width=\linewidth, trim={0 14cm 1cm 0cm}, clip]
    {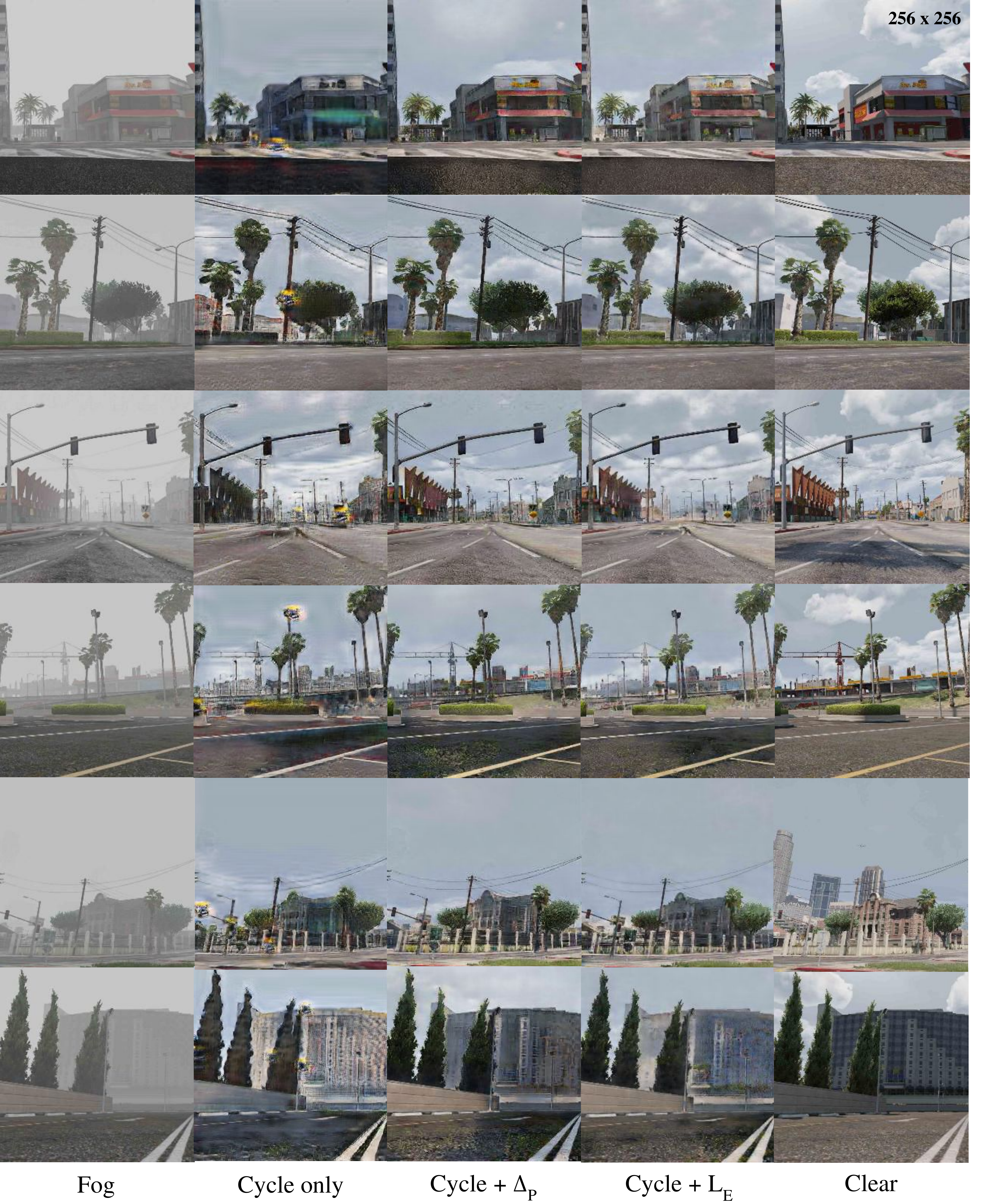}
   \caption{Different approaches for removing fog, part 1/2}
\label{fig:rem_fog_p1}
\end{figure*}

\begin{figure*}[t]
\centering
  \includegraphics[width=\linewidth, trim={0 13.75cm 1cm 0cm}, clip]
    {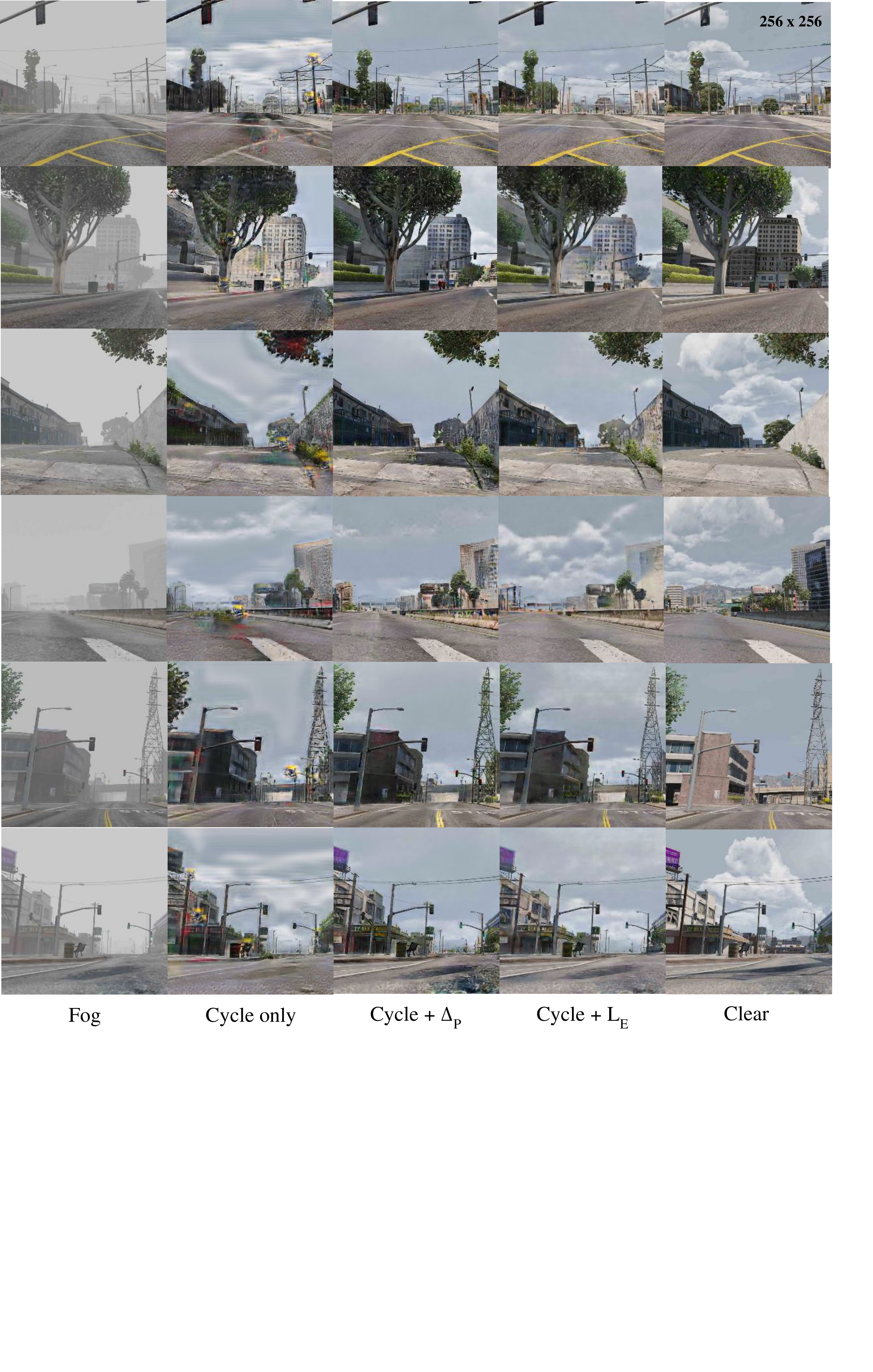}
   \caption{Different approaches for removing fog, part 2/2}
\label{fig:rem_fog_p2}
\end{figure*}

\begin{figure*}[t]
\centering
  \includegraphics[width=\linewidth, trim={0 47cm 4.5cm 0cm}, clip]
    {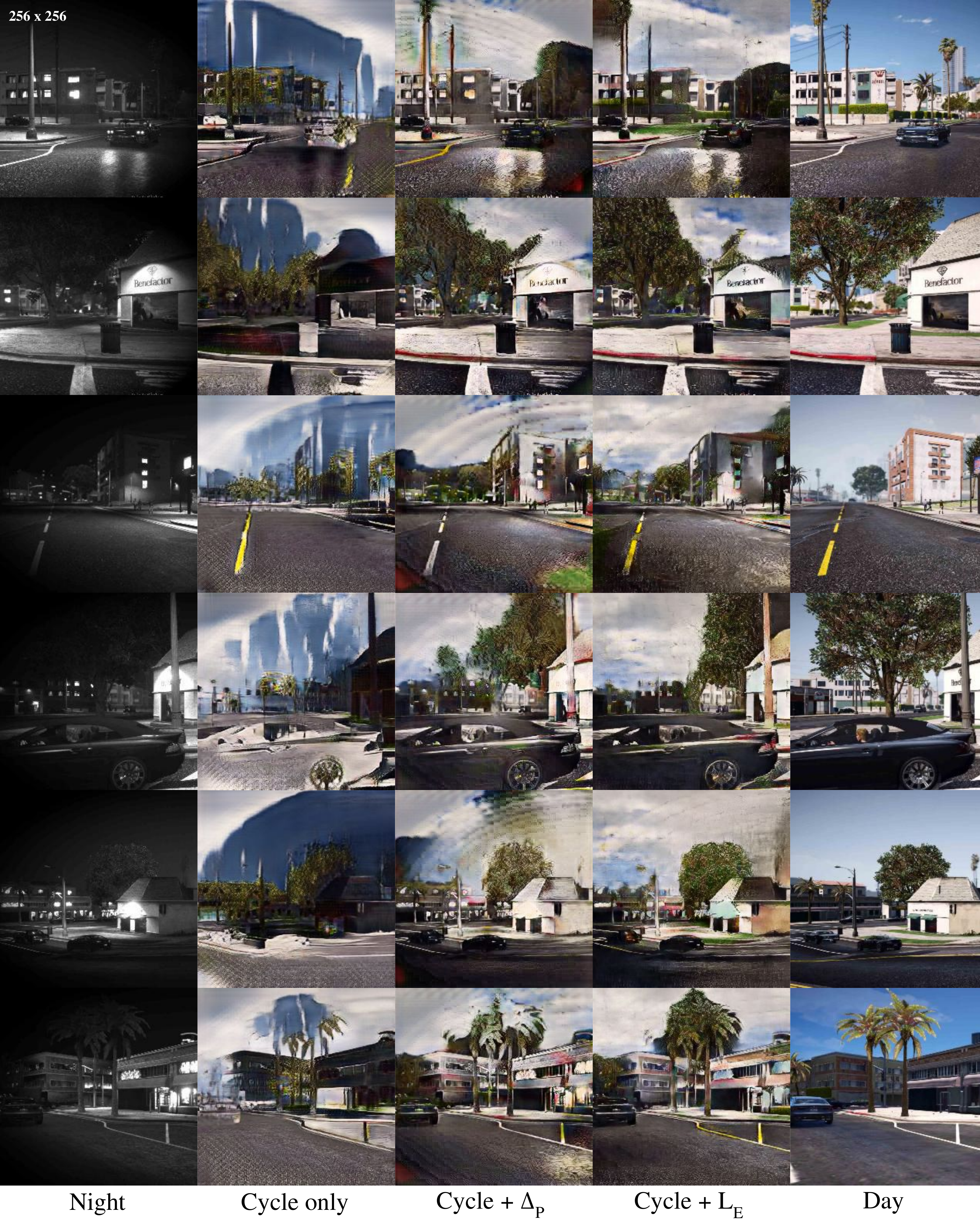}
   \caption{Different approaches for the night to day transformation on synthetically generated images, part 1/2}
\label{fig:n2d_synth_p1}
\end{figure*}

The results in the paper are presented in a smaller resolution due to space constraints. 
In this section we provide extra high resolution results for easier inspection.
For each task the $cycle$, $cycle+\Delta_{P}$ and $cycle+edge$ are compared side-by-side with the original image and for the synthetic datasets the corresponding image in the target domain is also shown.
The results for the night to day task are shown in Figure~\ref{fig:n2d_p1},~\ref{fig:n2d_p2},~\ref{fig:n2d_p3}~and~\ref{fig:n2d_p4} for the dataset with real images and in Figure~\ref{fig:n2d_synth_p1}~and~\ref{fig:n2d_synth_p2} for the dataset with synthetically generated images. In Figure~\ref{fig:rem_fog_p1}~and~\ref{fig:rem_fog_p2} the transformed images for the removing fog application are compared. 
The results in the figures are the complete images processed in one time. Although the transformers are trained with slightly smaller crops, the results do not degrade a lot.

\begin{table*}
   \footnotesize
    \centering
    \caption{Overview of the parameters}
    \label{tbl:overview_params}
	\begin{tabular}{cl}
        Symbol/Abbrev.		& Description\\
        \hline
        $\lambda_{cy, A}$  &	weight for the cycle consistency loss between an original image from domain A and its reconstruction. (a $->$ fb $->$ ra)   \\
        $\lambda_{cy, B}$   & weight for the cycle consistency loss between an original image from domain B and its reconstruction. (b $->$ fa $->$ rb)   \\
        
        $\lambda_{p, afb}$  & weight for the perceptual loss between an original image from domain A and its transformed fake image in domain B \\
        $\lambda_{p, bfa}$  & weight for the perceptual loss between an original image from domain B and its transformed fake image in domain A \\
        $\lambda_{p, farb}$  & weight for the perceptual loss between a transformed fake image in domain B and the reconstructed image in domain A \\
        $\lambda_{p, fbra}$  & weight for the perceptual loss between a transformed fake image in domain A and the reconstructed image in domain B \\
        
        $\lambda_{ep, afb}$  & weight for the edge preservation loss between an original image from domain A and its transformed fake image in domain B \\
        $\lambda_{ep, bfa}$  & weight for the edge preservation loss between an original image from domain B and its transformed fake image in domain A \\
        $\lambda_{ep, farb}$  & weight for the edge preservation loss between a transformed fake image in domain B and the reconstructed image in domain A \\
        $\lambda_{ep, farb}$  & weight for the edge preservation loss between a transformed fake image in domain A and the reconstructed image in domain B \\
        
        $\lambda_{ei, afb}$  & weight for the edge introduction loss between an original image from domain A and its transformed fake image in domain B \\
        $\lambda_{ei, bfa}$  & weight for the edge introduction loss between an original image from domain B and its transformed fake image in domain A \\
        $\lambda_{ei, farb}$  & weight for the edge introduction loss between a transformed fake image in domain B and the reconstructed image in domain A \\
        $\lambda_{ei, fbra}$  & weight for the edge introduction loss between a transformed fake image in domain A and the reconstructed image in domain B \\
        
        Load size  & input images are scaled to this size \\
        Crop size & the size of the random crop taken out of the resized image during training\\
        
        $N_{iter}$  & Number of iterations at start learning rate \\
        $N_{iter, decay}$  & Number of iterations to linearly decay learning rate to zero \\
        
        $Lr$  & Initial learning rate for adam \\
        $\beta_{1}$  & Momentum term of adam \\
        
        Pool size  & The size of image buffer that stores previously generated images for the discriminator \\

    \end{tabular}
\end{table*}

\begin{figure*}[t]
\centering
  \includegraphics[width=\linewidth, trim={0 1.5cm 4.5cm 45.5cm}, clip]
    {gta_n2d_large_tile_backup.pdf}
   \caption{Different approaches for the night to day transformation on synthetically generated images, part 2/2}
\label{fig:n2d_synth_p2}
\end{figure*}

\end{document}